\documentclass[conference]{IEEEtran}
\IEEEoverridecommandlockouts
\usepackage{cite}
\usepackage{amsmath,amssymb,amsfonts}

\usepackage{algorithm}  
\usepackage{algorithmicx}  
\usepackage[noend]{algpseudocode}

\usepackage{graphicx}
\usepackage{subcaption} 
\usepackage{array}

\usepackage{booktabs} 
\usepackage{multirow} 

\usepackage{cite}

\usepackage{textcomp}
\usepackage{xcolor}
\def\BibTeX{{\rm B\kern-.05em{\sc i\kern-.025em b}\kern-.08em
    T\kern-.1667em\lower.7ex\hbox{E}\kern-.125emX}}
\begin{document}

\title{Towards Initialization-Agnostic Clustering with Iterative Adaptive Resonance Theory
}
\author{
        \IEEEauthorblockN{ 
        Xiaozheng Qu\textsuperscript{1},
        Zhaochuan Li\textsuperscript{2},
        Zhuang Qi\textsuperscript{1},
        Xiang Li\textsuperscript{1},
        Haibei Huang\textsuperscript{2},
        Lei Meng\IEEEauthorrefmark{1}\textsuperscript{1},
        and Xiangxu Meng\textsuperscript{1}
        }
    \IEEEauthorblockA{\textsuperscript{1} School of Software, Shandong University, Jinan, China\\}
    \IEEEauthorblockA{\textsuperscript{2} Inspur, China\\}
    
    Email: \{xiaozhengqu, z\_qi, xiangli\_\}@mail.sdu.edu.cn, \{lizhaoch, huanghaibei\}@inspur.com, \{lmeng, mxx\}@sdu.edu.cn

}

\maketitle

\begin{abstract}
The clustering performance of Fuzzy Adaptive Resonance Theory (Fuzzy ART) is highly dependent on the preset vigilance parameter, where deviations in its value can lead to significant fluctuations in clustering results, severely limiting its practicality for non-expert users. Existing approaches generally enhance vigilance parameter robustness through adaptive mechanisms such as particle swarm optimization and fuzzy logic rules. However, they often introduce additional hyperparameters or complex frameworks that contradict the original simplicity of the algorithm. To address this, we propose Iterative Refinement Fuzzy Adaptive Resonance Theory (IR-ART), which integrates three key phases into a unified iterative framework: (1) \textit{Cluster Stability Detection}: A dynamic stability detection module that identifies unstable clusters by analyzing the change of sample size (number of samples in the cluster) in iteration. (2) \textit{Unstable Cluster Deletion}: An evolutionary pruning module that eliminates low-quality clusters. (3) \textit{Vigilance Region Expansion}: A vigilance region expansion mechanism that adaptively adjusts similarity thresholds. Independent of the specific execution of clustering, these three phases sequentially focus on analyzing the implicit knowledge within the iterative process, adjusting weights and vigilance parameters, thereby laying a foundation for the next iteration. Experimental evaluation demonstrates that IR-ART improves tolerance to suboptimal vigilance parameter values while preserving the parameter simplicity of Fuzzy ART. Case studies visually confirm the algorithm's self-optimization capability through iterative refinement, making it particularly suitable for non-expert users in resource-constrained scenarios.
\end{abstract}


\section{Introduction}
\noindent The performance of Fuzzy Adaptive Resonance Theory (Fuzzy ART) \cite{carpenter1991fuzzy} largely depends on the vigilance parameter \(\rho\), which is predetermined and remains constant throughout the process. A larger \(\rho\) results in a wider vigilance region (VR), imposing stricter similarity requirements on samples, and is more likely to generate numerous fine clusters. In contrast, a smaller \(\rho\) leads to a narrower VR, allowing data points to be included in the same cluster under more relaxed criteria. However, the traditional approach employs a uniform \(\rho\) globally and keeps it unchanged, making the algorithm highly sensitive to variations in \(\rho\). This dependency significantly increases the requirement for user expertise, complicating the attainment of ideal clustering results. Therefore, the ability to adaptively adjust \(\rho\) is crucial for balancing the trade-off between overfitting and underfitting.

Recent adaptive improvements to address the sensitivity issue of the vigilance parameter in the ART model include notable contributions such as combining overlapped categories and enhancing classification accuracy \cite{isawa2008fuzzy}. Methods for variable vigilance parameters and vigilance adaptation were introduced, refining the model’s dynamic responsiveness \cite{isawa2009fuzzy}\cite{meng2013vigilance}. Techniques for adaptive scaling of cluster boundaries specifically targeting large-scale social media data clustering were proposed to manage diverse data representations \cite{meng2015adaptive}. Particle swarm optimization techniques were implemented to dynamically adjust vigilance values, enhancing clustering effectiveness\cite{smith2015particle}. Most recently, salience-aware approaches for adaptive resonance theory improved clustering performance on sparse data \cite{meng2019salience}. Despite these advancements, a high usage threshold remains for non-expert users, highlighting the need for algorithms that balance enhanced capabilities with simplicity and usability.

To enhance the vigilance parameter robustness of Fuzzy ART while ensuring its efficiency and ease of understanding, thereby reducing the difficulty of use, we propose the Iterative Refinement Fuzzy Adaptive Resonance Theory (IR-ART). 
Building upon the traditional iterative process of Fuzzy ART, IR-ART introduces three core phases to form a new Iterations module, which includes: Cluster Stability Detection (CSD), Unstable Clusters Deletion (UCD), and Vigilance Region Expansion (VRE). These three phases do not perform clustering activities but instead focus on analyzing and adjusting the cluster weights and vigilance parameters, aiming to lay a stronger foundation for the next iteration. The CSD phase utilizes the knowledge about the sample size (number of samples in the cluster) acquired from the previous and current iterations to evaluate the stability of current clusters. Subsequently, the UCD phase removes clusters identified as unstable in the CSD phase, while the VRE phase fine-tunes \( \rho \) to enhance the influence of stable clusters. Through this iterative process, IR-ART achieves an enhanced robustness of $\rho$ without introducing additional parameters, preserving Fuzzy ART's operational simplicity while enabling non-expert users to obtain stable clustering across diverse vigilance settings.

We conducted experiments across 15 datasets, evaluating our method's effectiveness and universality in terms of Peak Performance, Mean Performance, and Standard Deviation. We also plotted clustering metrics as functions of $\rho$ to demonstrate IR-ART's performance improvement and robustness to $\rho$ variations. Finally, we visualized the execution process of IR-ART, showcasing its advantages through a real case.


The main contributions of this paper are as follows:
\begin{itemize}
    \item We propose a novel cluster stability detection method that identifies unstable clusters by analyzing dynamic changes in sample assignment during iteration. This replaces traditional complex computations and manual intervention.
    
    \item We design a strategy that jointly deletes unstable clusters and adjusts the vigilance parameter, enabling its self-adaptation and improving robustness without introducing additional predefined parameters.
\end{itemize}


\section{Related Work}

\noindent Adaptive Resonance Theory (ART) has evolved significantly since its introduction, addressing various challenges in machine learning and data clustering. The early foundational contributions established the theoretical basis, with ART 2 enabling stable category recognition for analog input patterns \cite{carpenter1987art}, and ART 2-A improving rapid category learning \cite{carpenter1991art2}. Fuzzy ART introduced fuzzy logic to enhance stability in learning and categorization of analog patterns \cite{carpenter1991fuzzy}, while ARTMAP extended the framework to support real-time supervised classification of dynamic data \cite{CARPENTER1991ARTMAP}. The distributed ARTMAP further improved the scalability for fast distributed learning \cite{carpenter1998distributed}, forming the basis for many subsequent innovations.

To address the sensitivity of ART to data variability, researchers introduced mechanisms such as dynamic vigilance adaptation to improve adaptability in non-stationary environments \cite{meng2013vigilance} and scaling of cluster boundaries for handling large-scale clustering tasks \cite{meng2015adaptive}. Interventions like interval type-2 fuzzy logic \cite{majeed2018uncertain} and validity index-guided vigilance tests \cite{da2017validity} enhanced ART’s stability and performance. Additionally, the mitigation of ordering effects using visualization techniques like VAT \cite{da2018study} contributed to improving clustering robustness. Information-theoretic approaches \cite{da2016information}, topological clustering \cite{masuyama2019topological}, and kernel Bayesian techniques \cite{masuyama2019kernel} further advanced the ability of ART to handle complex data structures. Innovations such as dual vigilance models \cite{da2019dual} and hypersphere-based cluster representations \cite{elnabarawy2019dual} provided finer control over clustering granularity and shape.

Recent developments have expanded ART’s scalability and flexibility. Distributed frameworks now support online learning while mitigating sensitivity to presentation order \cite{da2020distributed}. Hierarchical clustering algorithms \cite{tashiro2024growing}, biclustering methods for relational data \cite{yelugam2023topological}, and salience-aware clustering techniques \cite{meng2019salience} have broadened its applicability. iCVI-ARTMAP, incorporating incremental cluster validity indices, has demonstrated improvements in validation efficiency and robustness to data order \cite{da2022icvi}\cite{da2022incremental}. These advancements underscore the ongoing relevance and adaptability of ART to modern data analysis.

\section{Fuzzy ART}

\noindent In Fuzzy Adaptive Resonance Theory (Fuzzy ART), $F_1$ represents the input field, which receives the input samples, and $F_2$ represents the category field, which stores the cluster information. The main content of Fuzzy ART is as follows.

\textbf{Input vectors:} Let $\mathbf{x}$ denote a sample in the feature space, where $ \mathbf{x} = (x_1, \ldots, x_m) $ and \( x_i \in [0, 1] \) for \( i = 1, \ldots, m \), and \( m \) represents the dimension of the sample. With the complement coding operation, \( \mathbf{x} \) is concatenated with its complement vector \( \mathbf{\bar{x}} \) to form the final input feature vector \( \mathbf{I} = (\mathbf{x}, \mathbf{\bar{x}}) \) in the input field \( F_1 \), where \( \mathbf{\bar{x}} = 1 - \mathbf{x} \).

\textbf{Weight Vector:} Let \( \mathbf w_j \) represent the weight vector of the \( j \)-th cluster \( C_j \) (\( j = 1, \ldots, J \)) in the category field \( F_2 \).

\textbf{Parameters:} The key parameters in the Fuzzy ART algorithm include the choice parameter \( \alpha > 0 \), the learning parameter \( \beta \in [0, 1] \), and the vigilance parameter \( \rho \in [0, 1] \). In practice, $\alpha$ is typically chosen to be a very small value.

The three key steps of Fuzzy ART are as follows:

\begin{enumerate}
\item \textbf{Category choice:} For each input vector $\mathbf{I}$ in $F_1$, Fuzzy ART calculates the choice function $T_j$ for each cluster $C_j$ in $F_2$, and selects the cluster $C_{j*}$ as the winner cluster, where \( j^* = \arg\max_j T_j \).
The choice function $T_j$ for $\mathbf{I}$ and $C_j$ is computed as follows:
\begin{equation}
\label{eq:Choice Function}
T_j = \frac{\left | \mathbf{I} \land \mathbf{w}_j  \right |}{\alpha + \left | \mathbf{w}_j \right |},
\end{equation}
where $\left| \cdot \right|$ denotes the $L_1$ norm, and operation $\land$ is defined by $(\mathbf{p} \land \mathbf{q})_i \equiv \min(p_i, q_i)$.

\item \textbf{Template matching:}
For $\mathbf{I}$ and the winner cluster $C_{j^*}$, the match function is calculated as follows:
\begin{equation}
M_{j^*} = \frac{\left| \mathbf{I} \wedge \mathbf{w}_{j^*} \right|}{\left| \mathbf{I} \right|}.
\end{equation}
If $M_{j^*} \geq \rho$, resonance occurs and the input vector $\mathbf{I}$ (representing the sample \(\mathbf{x}\)) is assigned to the winner cluster $C_{j^*}$, followed by weight vector updating in Step 3. If $M_{j^*} < \rho$, a new winner cluster is selected from the remaining clusters in $F_2$. If no winner cluster satisfies $\rho$, a new cluster is created to encode $\mathbf{I}$.

\item \textbf{Prototype learning:} If resonance occurs, the weight vector \( \mathbf{w}_{j^*} \) is updated as follows:
    \begin{equation}
    \mathbf{w}_{j^*}^{(new)} = \beta (\mathbf{I} \wedge \mathbf{w}_{j^*}) + (1 - \beta)\mathbf{w}_{j^*}.
    \end{equation}
\end{enumerate}

\vspace{-0.1cm}

\section{Iterative Refinement Fuzzy ART}

\subsection{Overall framework}

\begin{figure*}[t]   
  \includegraphics[width=0.99\textwidth]{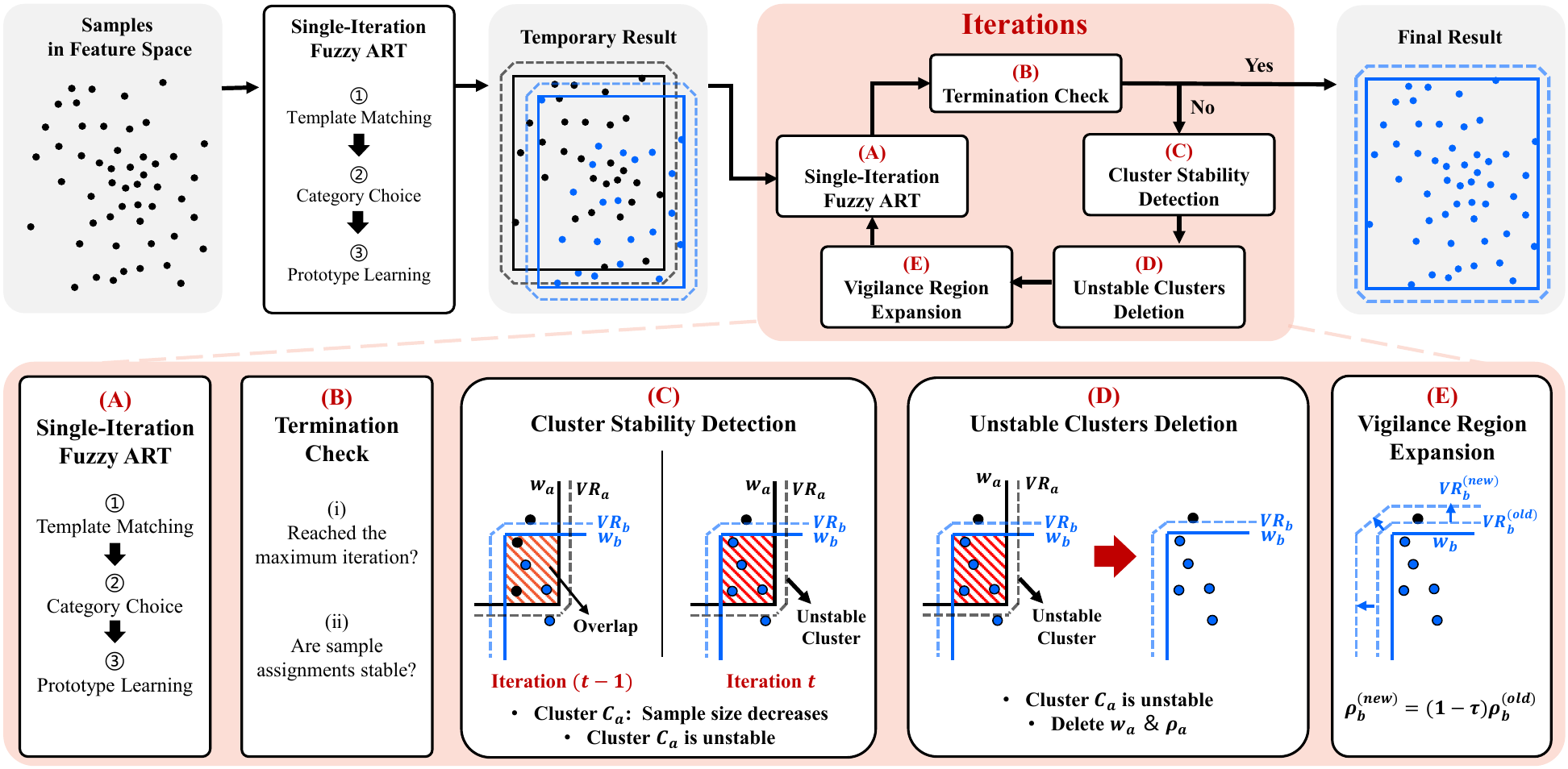}
  \centering
  \caption{
  The upper part of this figure briefly presents the overall framework of IR-ART and the Iterations module is detailed in the lower part of this figure. Phases (A) and (B) in the Iterations are similar to those in traditional Fuzzy ART iterations, whereas the core phases of IR-ART are (C), (D), and (E).
  }
  \label{fig:framework}
  \vspace{-0.5cm}
\end{figure*}
Fuzzy ART is known to refine weight vectors and improve clustering performance by repeating all samples in the dataset. We optimize this iterative process and propose a novel algorithm called Iterative Refinement Fuzzy Adaptive Resonance Theory (IR-ART). The overall framework of IR-ART is illustrated in Fig.~\ref{fig:framework}, and its pseudocode is provided in Algorithm~\ref{alg:pseudocode}. For clarity, we denote the \(t\)-th iteration of our method as Iteration \(t\). IR-ART begins with a single Fuzzy ART execution on all samples, detailed in lines 1-18 of the pseudocode. This completes Iteration 1, resulting in a Temporary Result that serves as the basis for the subsequent \textbf{Iterations} module. This module comprises the following five phases, among which (A) and (B) inherit the iterative steps of Fuzzy ART:
\begin{enumerate}
\item[(A)] \textbf{Single-Iteration Fuzzy ART:} This phase marks the beginning of the Iterations module. In this phase, the iteration count \(t\) is increased by 1, and a single execution of the Fuzzy ART algorithm is performed on all samples to update the existing weight vectors and sample assignments before proceeding to the next phase.
\item[(B)] \textbf{Termination Check:} This phase evaluates whether the algorithm satisfies the predefined termination criteria. If the maximum iteration count is reached or stable sample assignments are achieved, the algorithm terminates and outputs the current assignments as the Final Result.
\item[(C)] \textbf{Cluster Stability Detection (CSD):} This phase employs a heuristic method to compare the current and previous sample assignment results, identifying stable and unstable clusters. This step lays the foundation for the subsequent two phases.
\item[(D)] \textbf{Unstable Clusters Deletion (UCD):} The weight vectors corresponding to the unstable clusters identified during the CSD phase are removed in this phase.
\item[(E)] \textbf{Vigilance Region Expansion (VRE):} The vigilance parameters of the retained clusters are slightly adjusted, increasing the likelihood of input vectors being incorporated into these clusters in the next iteration.
\end{enumerate}
\vspace{-0.1cm}

As shown in Fig.~\ref{fig:framework}, phases (A) to (E) in the Iterations module are executed sequentially through multiple iterations, corresponding to lines 19-33 in the pseudocode. We set \(\alpha = 0.001\), following \cite{da2020distributed}, and prior studies show clustering performance is generally robust to this parameter \cite{tan1995adaptive}. \(\tau\) is given a default small positive value, and \(t_{\text{max}}\) can be flexibly set to limit the number of iterations. Therefore, the usage of our method remains as simple as Fuzzy ART, with clustering performance primarily controlled by the preset vigilance parameter \(\rho\), which constrains intra-cluster similarity and has a much greater impact on results than the learning rate \(\beta\).

Subsequent sections will dive into the three core phases of our method: CSD, UCD, and VRE.

\renewcommand{\algorithmicrequire}{\textbf{Input:}}
\renewcommand{\algorithmicensure}{\textbf{Output:}}
\begin{algorithm}[t]
\caption{IR-ART}
\label{alg:pseudocode}
\small 
\begin{algorithmic}[1]
\Require
    Set of input vectors $S=\{\mathbf{I}_1,...,\mathbf{I}_N\}$, vigilance parameter $\rho_0$, learning rate $\beta$, choice parameter $\alpha=0.001$, expansion parameter $\tau = 0.01$, maximum number of iterations $t_{\text{max}}$.
\Ensure
    Set of clusters $K$, set of weight vectors $W$, sample assignment $G_t$ in Iteration \(t\).

\State Initialize $K=\emptyset$, $W=\emptyset$.
\State Initialize current number of iterations $t=1$ and initialize $G_t$.
\State Create cluster $C_1$ using the first sample $\mathbf{I}_1$ and $\rho_0$.
\State Update $K$ and $W$, update $G_1[1] = 1$.
\For{each input vector $\mathbf{I}_n$ $(n = 2, \ldots, N)$}
    \For{each cluster $C_j$ in $K$}
        \State Calculate the choice function $T_j$.
    \EndFor
    \State Select the winner cluster index $j^* = \arg\max_j T_j$.
    \State Calculate the match function $M_{j^*}$.
    \If{$M_{j^*} < \rho_{j^*}$}
        \State Select a new winner cluster index $j^*$ and go to step \textbf{9}.
    \Else
        \State Update $\mathbf{w}_{j^*}$ in $W$.
        \State Update $G_t[n] = j^*$.
    \EndIf
    \If{no winner satisfies the vigilance parameter}
        \State Create a new cluster $C_{(new)}$ using $\mathbf{I}_n$ and $\rho_0$.
        \State Update $K$ and $W$.
        \State Update $G_t[n] = \text{new cluster index}$.
    \EndIf
\EndFor
\While{true}
    \State Set $t = t + 1$.
    \For{each input vector $\mathbf{I}_n$ $(n = 1, \ldots, N)$}
        \State Perform Fuzzy ART, the same as steps \textbf{6} to \textbf{18}.
    \EndFor
    \If{$t = t_{\text{max}}$ or $G_t$ = $G_{t-1}$}
        \State \Return $K$, $W$ and $G_t$.
    \Else
        \State Initialize unstable cluster set $U = \emptyset$.
        \State Compare $G_t$ and $G_{t-1}$ to identify clusters with a reduced
        \Statex \hspace{3em}sample size and add them to $U$.
        \For{each cluster $C_j$ in $K$}
            \If{$C_j$ is in $U$}
                \State Delete $C_j$ from $K$ and delete $\mathbf{w}_j$ from $W$.
                \State Update $G_t$.
            \EndIf
        \EndFor
        \For{each remaining cluster $C_j$ in $K$}
            \State Set $\rho_j^{(new)} = (1 - \tau) \rho_j^{(old)}$.
        \EndFor
    \EndIf
\EndWhile
\end{algorithmic}
\end{algorithm}

\subsection{Cluster Stability Detection}
In this section, we provide a detailed explanation of the Cluster Stability Detection (CSD) phase, as shown in (C) of Fig.~\ref{fig:framework} and lines 26–27 of the pseudocode.

As previously mentioned, repeatedly processing all input samples typically improves the clustering performance of Fuzzy ART, largely due to the gradual optimization of weight vectors. This allows some samples to be reassigned to more suitable clusters. As a result, a cluster \(C_j\) in category field \(F_2\) may gain or lose samples over iterations, resulting in changes in its number of contained samples (referred to as sample size). Based on this observation, we designed a simple and intuitive method to take advantage of these changes during the iterative process to assess cluster stability. 

Specifically, let \(G_t\) represent the sample assignment in Iteration \(t\), recording the cluster index for each sample. During CSD, the algorithm compares the sample sizes of each cluster between Iteration \(t\) and \(t{-}1\), identifying clusters with decreased size as unstable. Clusters with unchanged or increased size, including new ones in Iteration \(t\), are considered stable.

The rationale is that clusters with reduced sample sizes are more likely to have imperfect weight vectors or \(\rho\), leading to unstable cluster structures that make them more prone to losing samples during iterations. In contrast, clusters with stable or increased sample sizes are considered to form more meaningful cluster structures, which provide them with an advantage during the iterative process.

As shown in (C) of Fig.~\ref{fig:framework}, the clusters \(C_a\) and \(C_b\) have an overlapping vigilance region, which is a typical scenario in which changes in the sample size of the clusters are likely to occur. At the end of Iteration \(t{-}1\), some samples lie within the overlap. In Iteration \(t\), after one execution of Fuzzy ART, some samples of \(C_a\) are “captured” by \(C_b\), resulting in a decrease in the sample size of \(C_a\). Therefore, \(C_a\) is identified as unstable, while \(C_b\) is considered stable.

\subsection{Unstable Clusters Deletion}
The Unstable Clusters Deletion (UCD) phase is depicted in (D) of Fig.~\ref{fig:framework} and lines 28–31 of the pseudocode.

Based on the results obtained during the CSD phase, the UCD phase iterates through all clusters in the current category field \(F_2\). If a cluster is identified as unstable during the CSD phase, it is removed from \(F_2\), preventing its influence on the Fuzzy ART clustering process in the next iteration. This involves deleting the corresponding weight vector and the vigilance parameter and updating $G_t$. As shown in Fig.~\ref{fig:framework}(D), \(\mathbf{w}_a\) and \(\rho_a\) of the unstable cluster \(C_a\) identified in the previous phase are removed, which is reflected in the figure as the disappearance of the weight rectangle and the \(\mathrm{VR}_a\).

The rationale behind this deletion mechanism is straightforward: survival of the fittest. Specifically, unstable clusters are considered to have imperfect weight vectors or \(\rho\), which may lead to competition with stable clusters for sample assignments during the iterative process, thereby hindering improvements in clustering performance. By removing unstable clusters, they are excluded from competing during the next iteration of the Fuzzy ART clustering process. This minimizes the influence of unstable clusters and ensures that high-quality clusters are prioritized in subsequent iterations, ultimately contributing to improved clustering outcomes.

\subsection{Vigilance Region Expansion}
The Vigilance Region Expansion (VRE) phase is depicted in (E) of Fig.~\ref{fig:framework} and lines 32–33 of the pseudocode.

In Fuzzy ART, the vigilance parameter \(\rho\) determines the similarity threshold required for a sample to be assigned to a cluster. This forms a hyper-octagon in feature space, centered around the weight hyper-rectangle, known as the vigilance region (VR). A sample is eligible to be incorporated into the corresponding cluster only if it falls within this VR.

After the CSD and UCD phases, all clusters in the current category field \(F_2\) are stable. Based on the principle of survival of the fittest, we propose that a slight expansion of the VRs for these stable clusters can slightly enlarge the regions capable of absorbing new samples without disrupting the overall structure. This enhances the influence of high-quality clusters in subsequent iterations. Specifically, during the VRE phase, the algorithm adjusts the vigilance parameter \(\rho_j\) of each existing cluster \(C_j\) in \(F_2\) using the following formula:
\begin{equation}
    \rho_j^{(new)} = (1 - \tau) \rho_j^{(old)},  
\end{equation}
where \(\tau\) is a small positive constant controlling the expansion rate. For simplicity and general applicability, We set \(\tau = 0.01\) by default to ensure a moderate and stable adjustment.

By incorporating the VRE phase, our method gains the ability to dynamically adjust \(\rho\), thereby improving clustering performance even when \(\rho\) is suboptimal. Furthermore, the combination of VRE and UCD enables a complete processing of all clusters in \(F_2\) during the current iteration. This ensures that samples in the next iteration are more likely to be assigned to appropriate clusters rather than forming new clusters, particularly for those samples left unassigned after the UCD phase.


\section{Experiments}

\subsection{Datasets}
To evaluate the applicability of our algorithm in diverse datasets, we adopted the dataset selection strategy proposed in \cite{da2020distributed}. Our experiments involve a combination of 15 real world and artificial benchmark datasets with various characteristics, which are accessible from \cite{Bache+Lichman:2013,ClusteringDatasets,ilc2011gravitational}. The key properties of these datasets are summarized in Table~\ref{tab:datasets}. Before the experiments, all datasets were preprocessed using Min-Max normalization to scale their characteristic values to the range \([0, 1]\) and further transformed into input vectors through complement coding \cite{da2020distributed}.

\subsection{Experimental Setup}

Our experiments aim to simulate the clustering performance of a user lacking prior knowledge when employing ART-based clustering methods. We compared our method with Fuzzy ART\cite{carpenter1991fuzzy}, and four methods mentioned in Introduction: CM-ART\cite{meng2013vigilance}, AM-ART\cite{meng2013vigilance}, HI-ART\cite{meng2015adaptive}, and SA-ART\cite{meng2019salience}. These methods were selected for their minimal parameter requirements and adaptive mechanisms for the vigilance parameter.

For parameter settings, we avoided complex optimization for individual datasets. Instead, we applied a uniform set of reasonable and commonly used parameters across all datasets to minimize dependence on prior knowledge and ensure a fair comparison. Specifically, the choice parameter \(\alpha\) was set to \(10^{-3}\). The learning rate $\beta$ was set to 0.5, reflecting a moderate level of learning. In SA-ART, \(\lambda = 0.5\) balanced frequency and stability. For AM-ART, we evaluated $\sigma$ from the set $\{10^{-1}, 10^{-2}, 10^{-3}, 10^{-4}, 10^{-5}\}$ and selected $10^{-4}$, as it yielded reasonable cluster numbers in most datasets. Similarly, we set $\delta$ in SA-ART and $\Delta$ in CM-ART to $10^{-4}$. Other methods were iterated with $t_{\text{max}} = 50$, following the termination criteria specified in line 23 of the pseudocode, except for CM-ART and HI-ART, which were executed only once due to rapid cluster growth with further iterations. The initial vigilance parameter $\rho$ was the sole variable, scanned in the range $[0.05, 0.95]$ with a step size of 0.01. Each setting was run 10 times on randomly ordered datasets, and the average performance metrics corresponding to each \(\rho\) were recorded.

We employ two commonly used external cluster validity indices, Normalized Mutual Information (NMI) and Adjusted Rand Index (ARI), to evaluate the quality of the clustering results. Higher values of these indices generally indicate better clustering performance. 
More details can be found in \cite{zhou2024comprehensive}.


\begin{table}[t]
\renewcommand\arraystretch{0.6}
    \caption{Summary of the datasets}
    \centering
    \resizebox{\linewidth}{!}{
        \begin{tabular}{ccccc}
            \toprule
            dataset & samples & features & clusters & type \\
            \midrule
            \midrule
            Aggregation & 788 & 2 & 7 & Artificial \\
            Compound & 399 & 2 & 6 & Artificial \\
            Dermatology & 358 & 34 & 6 & Real World \\
            Face & 320 & 2 & 4 & Artificial \\
            Flag & 640 & 2 & 3 & Artificial \\
            Flame & 240 & 2 & 2 & Artificial \\
            Glass & 214 & 10 & 6 & Real World \\
            Iris & 150 & 4 & 3 & Real World \\
            Jain & 373 & 2 & 2 & Artificial \\
            Moon & 514 & 2 & 4 & Artificial \\
            Path based & 300 & 2 & 3 & Artificial \\
            Seeds & 210 & 7 & 3 & Real World \\
            Spiral & 312 & 2 & 3 & Artificial \\
            Synthetic Control & 600 & 60 & 6 & Real World \\
            Wave & 287 & 2 & 2 & Artificial \\
            \bottomrule 
        \end{tabular}
    }
    \label{tab:datasets}
    \vspace{-0.4cm}
\end{table}

\subsection{Clustering Performance Comparison}
As described in the previous section, for each value of $\rho$, experiments were conducted with 10 random data input orders. The average Normalized Mutual Information (aNMI) and the average Adjusted Rand Index (aARI) were calculated across these 10 trials. These two metrics are considered to represent the clustering performance of a given method for a specific $\rho$ value on a given dataset.

\begin{table*}[tbp]
\renewcommand\arraystretch{0.5}
\setlength{\tabcolsep}{4pt} 
    \caption{Clustering performance of algorithms in terms of NMI and ARI. Abbreviations are used for some datasets and FA means Fuzzy ART. The optimal results are shown in bold.}
    \centering
    \resizebox{\textwidth}{!}{
    \normalsize 
    \begin{tabular}{c|c| cccccc | cccccc | cc}
        \toprule
        
        \multicolumn{2}{c}{\multirow{2}{*}{}} 
        & \multicolumn{6}{|c|}{\textbf{Peak Performance}} & \multicolumn{6}{c}{\textbf{Mean Performance}} & \multicolumn{2}{|c}{\textbf{Standard Deviation}} \\
        
        \cmidrule(r){3-8} \cmidrule(lr){9-14} \cmidrule(l){15-16}
        
        \multicolumn{2}{c|}{~} 
        & FA & CM-ART & AM-ART & HI-ART & SA-ART & Ours & FA & CM-ART & AM-ART & HI-ART & SA-ART & Ours & SA-ART & Ours \\
        
        \midrule
        \midrule
        
        \multirow{2}{*}{Agg}
        & ARI 
        & 0.613 & 0.354 & 0.469 & 0.357 & 0.707 & \textbf{0.779} 
        & 0.372 & 0.221 & 0.259 & 0.230 & 0.301 & \textbf{0.502}
        & 0.215 & \textbf{0.160}
        \\
        & NMI 
        & 0.725 & 0.595 & 0.607 & 0.618 & 0.780 & \textbf{0.849} 
        & 0.591 & 0.529 & 0.523 & 0.541 & 0.499 & \textbf{0.674}
        & 0.263 & \textbf{0.125}
        \\
        
        \midrule
        
        \multirow{2}{*}{Com}
        & ARI 
        & 0.638 & 0.466 & 0.555 & 0.436 & 0.755 & \textbf{0.757}
        & 0.366 & 0.235 & 0.296 & 0.236 & 0.333 & \textbf{0.499}
        & 0.269 & \textbf{0.170}
        \\
        & NMI 
        & 0.691 & 0.591 & 0.606 & 0.587 & 0.770 & \textbf{0.779} 
        & 0.536 & 0.472 & 0.509 & 0.473 & 0.448 & \textbf{0.619} 
        & 0.294 & \textbf{0.128}
        \\

        \midrule

        \multirow{2}{*}{Derm}
        & ARI 
        & 0.300 & 0.095 & 0.144 & 0.105 & 0.509 & \textbf{0.618}
        & 0.134 & 0.061 & 0.076 & 0.067 & 0.115 & \textbf{0.336} 
        & 0.172 & \textbf{0.141}
        \\
        & NMI 
        & 0.498 & 0.501 & 0.479 & 0.505 & 0.642 & \textbf{0.648} 
        & 0.453 & 0.415 & 0.417 & 0.421 & 0.224 & \textbf{0.515} 
        & 0.276 & \textbf{0.101}
        \\
        
        \midrule
        
        \multirow{2}{*}{Face}
        & ARI
        & 0.466 & 0.442 & 0.600 & 0.429 & 0.216 & \textbf{0.609}
        & 0.147 & 0.151 & 0.127 & 0.138 & 0.091 & \textbf{0.229} 
        & \textbf{0.070} & 0.182
        \\
        & NMI
        & 0.446 & 0.522 & 0.558 & 0.507 & 0.473 & \textbf{0.624}
        & 0.315 & 0.313 & 0.275 & 0.303 & 0.244 & \textbf{0.399}
        & 0.168 & \textbf{0.089}
        \\
        
        \midrule
        
        \multirow{2}{*}{Flag}
        & ARI 
        & 0.895 & 0.593 & 0.842 & 0.481 & 0.730 & \textbf{0.979}
        & 0.479 & 0.323 & 0.339 & 0.288 & 0.424 & \textbf{0.529}
        & 0.276 & \textbf{0.235}
        \\
        & NMI
        & 0.893 & 0.670 & 0.869 & 0.597 & 0.798 & \textbf{0.978}
        & 0.618 & 0.502 & 0.515 & 0.483 & 0.554 & \textbf{0.655}
        & 0.263 & \textbf{0.209}
        \\
        
        \midrule
        
        \multirow{2}{*}{Flame}
        & ARI
        & 0.298 & 0.128 & 0.339 & 0.132 & 0.454 & \textbf{0.456}
        & 0.162 & 0.060 & 0.135 & 0.062 & 0.176 & \textbf{0.215} 
        & 0.159 & \textbf{0.104}
        \\
        & NMI
        & 0.376 & 0.317 & 0.343 & 0.320 & \textbf{0.510} & 0.432
        & 0.290 & 0.240 & 0.269 & 0.241 & 0.245 & \textbf{0.324} 
        & 0.191 & \textbf{0.079}
        \\
        
        \midrule
        
        \multirow{2}{*}{Glass}
        & ARI
        & 0.243 & 0.190 & 0.147 & 0.182 & \textbf{0.390} & 0.281
        & 0.149 & 0.088 & 0.082 & 0.090 & 0.106 & \textbf{0.180} 
        & 0.139 & \textbf{0.080}
        \\
        & NMI
        & 0.499 & 0.510 & 0.469 & 0.516 & \textbf{0.584} & 0.545
        & \textbf{0.345} & 0.272 & 0.309 & 0.278 & 0.201 & 0.317 
        & 0.244 & \textbf{0.156}
        \\

        \midrule
        
        \multirow{2}{*}{Iris}
        & ARI
        & 0.606 & 0.493 & 0.574 & 0.493 & \textbf{0.707} & 0.701
        & 0.372 & 0.230 & 0.314 & 0.231 & 0.320 & \textbf{0.464}
        & 0.265 & \textbf{0.169}
        \\
        & NMI
        & 0.659 & 0.554 & 0.606 & 0.555 & \textbf{0.750} & 0.730
        & 0.522 & 0.383 & 0.485 & 0.384 & 0.401 & \textbf{0.585}
        & 0.301 & \textbf{0.132}
        \\
        
        \midrule
        
        \multirow{2}{*}{Jain}
        & ARI
        & 0.616 & 0.378 & 0.733 & 0.247 & 0.560 & \textbf{0.782}
        & 0.291 & 0.167 & 0.249 & 0.139 & 0.210 & \textbf{0.401}
        & \textbf{0.197} & 0.210
        \\
        & NMI
        & 0.533 & 0.367 & 0.604 & 0.315 & 0.499 & \textbf{0.668}
        & 0.384 & 0.279 & 0.339 & 0.266 & 0.286 & \textbf{0.445}
        & 0.192 & \textbf{0.089}
        \\

        \midrule
        
        \multirow{2}{*}{Moon}
        & ARI
        & 0.295 & 0.219 & 0.251 & 0.222 & 0.290 & \textbf{0.333}
        & 0.226 & 0.163 & 0.142 & 0.159 & 0.152 & \textbf{0.228}
        & 0.113 & \textbf{0.061}
        \\
        & NMI
        & 0.563 & 0.541 & 0.480 & 0.542 & 0.593 & \textbf{0.634}
        & 0.417 & 0.431 & 0.344 & \textbf{0.434} & 0.300 & 0.390
        & 0.227 & \textbf{0.135}
        \\
        
        \midrule
        
        \multirow{2}{*}{Path}
        & ARI
        & 0.333 & 0.307 & 0.281 & 0.299 & 0.454 & \textbf{0.468}
        & 0.198 & 0.150 & 0.152 & 0.151 & 0.210 & \textbf{0.272}
        & 0.185 & \textbf{0.115}
        \\
        & NMI
        & 0.464 & 0.446 & 0.406 & 0.446 & 0.525 & \textbf{0.536}
        & 0.344 & 0.346 & 0.321 & 0.347 & 0.294 & \textbf{0.383}
        & 0.234 & \textbf{0.123}
        \\
        
        \midrule
        
        \multirow{2}{*}{Seeds}
        & ARI
        & 0.482 & 0.265 & 0.459 & 0.279 & \textbf{0.651} & 0.633
        & 0.252 & 0.131 & 0.213 & 0.134 & 0.254 & \textbf{0.351}
        & 0.235 & \textbf{0.156}
        \\
        & NMI
        & 0.526 & 0.421 & 0.541 & 0.424 & \textbf{0.646} & 0.633
        & 0.422 & 0.323 & 0.406 & 0.324 & 0.329 & \textbf{0.480}
        & 0.260 & \textbf{0.094}
        \\
        
        \midrule
        
        \multirow{2}{*}{Spiral}
        & ARI
        & 0.098 & 0.091 & 0.052 & 0.093 & 0.120 & \textbf{0.127}
        & \textbf{0.052} & 0.046 & 0.026 & 0.045 & 0.024 & 0.037
        & 0.041 & \textbf{0.039}
        \\
        & NMI
        & 0.449 & 0.442 & 0.396 & 0.443 & \textbf{0.471} & 0.468
        & 0.166 & \textbf{0.243} & 0.153 & \textbf{0.243} & 0.083 & 0.117
        & 0.146 & \textbf{0.134}
        \\
        
        \midrule
        
        \multirow{2}{*}{Syn}
        & ARI
        & 0.066 & 0.105 & 0.096 & 0.179 & \textbf{0.578} & 0.506
        & 0.035 & 0.050 & 0.055 & 0.076 & 0.155 & \textbf{0.250}
        & 0.189 & \textbf{0.175}
        \\
        & NMI
        & 0.467 & 0.477 & 0.474 & 0.486 & \textbf{0.735} & 0.669 
        & 0.370 & 0.363 & 0.388 & 0.383 & 0.306 & \textbf{0.415}
        & 0.300 & \textbf{0.215}
        \\
        
        \midrule
        
        \multirow{2}{*}{Wave}
        & ARI
        & 0.170 & 0.133 & 0.119 & 0.133 & 0.153 & \textbf{0.250}
        & 0.101 & 0.081 & 0.064 & 0.079 & 0.060 & \textbf{0.120}
        & \textbf{0.049} & 0.059
        \\
        & NMI
        & 0.383 & 0.355 & 0.313 & 0.355 & 0.389 & \textbf{0.437} 
        & 0.210 & \textbf{0.262} & 0.174 & 0.261 & 0.122 & 0.213
        & 0.137 & \textbf{0.136}
        \\
        
        \bottomrule
    \end{tabular}
    }
    \label{tab:main_table}
    \vspace{-0.4cm}
\end{table*}

The maximum values of aARI and aNMI obtained during the $\rho$ scan were taken and are presented in the Peak Performance section of Table~\ref{tab:main_table}. This indicates the best clustering performance achieved by each algorithm during the scan. The peak performance of CM-ART, AM-ART, and HI-ART fluctuates across different datasets, and they do not consistently outperform Fuzzy ART. This is likely due to the distinct characteristics of the datasets and the imperfect settings of other predefined parameters besides $\rho$. This implies that achieving good clustering results with these methods often requires appropriate adjustments to predefined parameters based on different data scenarios, which places a higher demand on prior knowledge. SA-ART achieves the best peak performance on high-dimensional datasets, such as Glass and Synthetic Control, consistent with its known strengths~\cite{meng2019salience}. IR-ART achieves superior performance on lower-dimensional datasets such as Face and Flag, while also surpassing all other methods except SA-ART in peak performance on higher-dimensional datasets. This suggests that, without specific parameter tuning, SA-ART and IR-ART generally exhibit superior peak clustering ability compared to other methods. This also demonstrates the performance improvement of IR-ART over Fuzzy ART.

For ART-based algorithms, only a limited number of \( \rho\) values usually lead to peak performance. Users lacking prior knowledge are more likely to select a non-optimal \( \rho\) than an optimal one. So we separately calculated the averages of all aNMI and aARI values obtained during the \( \rho\) scan, referred to as mean average Normalized Mutual Information (mNMI) and mean aARI (mARI), to evaluate the general clustering performance of the algorithms. These results are presented in the Mean Performance section of Table~\ref{tab:main_table}. The results indicate that, whether measured by mNMI or mARI, IR-ART achieves the best performance on most datasets but performs poorly on a few with specific cluster shapes, such as Spiral. The other methods show varying strengths and weaknesses across different datasets, with SA-ART not standing out in general clustering performance. Furthermore, to investigate the fluctuations of aNMI and aARI during the scan, we calculated their standard deviations, denoted as sNMI and sARI, and compared them with those of SA-ART. These results are presented in the Standard Deviation section of Table~\ref{tab:main_table}. It indicates that IR-ART is less sensitive to changes in \( \rho \) during the scan, whereas SA-ART is more responsive, as sNMI and sARI are higher for SA-ART than for IR-ART across most datasets. In summary, we conclude that when users randomly select \( \rho \) due to a lack of prior knowledge, IR-ART is often the preferred choice among these methods, while other algorithms, such as SA-ART, are more susceptible to suboptimal \( \rho \), which can adversely affect their performance. This shows that IR-ART effectively improves general performance and mitigates the sensitivity of ART-based clustering algorithms to \( \rho \). Furthermore, in terms of volatility, IR-ART is more stable than SA-ART. This indicates that IR-ART not only achieves good overall performance but also ensures actual results remain close to expectations, thus alleviating the influence of \( \rho \).

\begin{figure*}[tbp]
    \centering
    \begin{subfigure}[b]{0.25\textwidth}
        \centering
        \includegraphics[width=\textwidth]{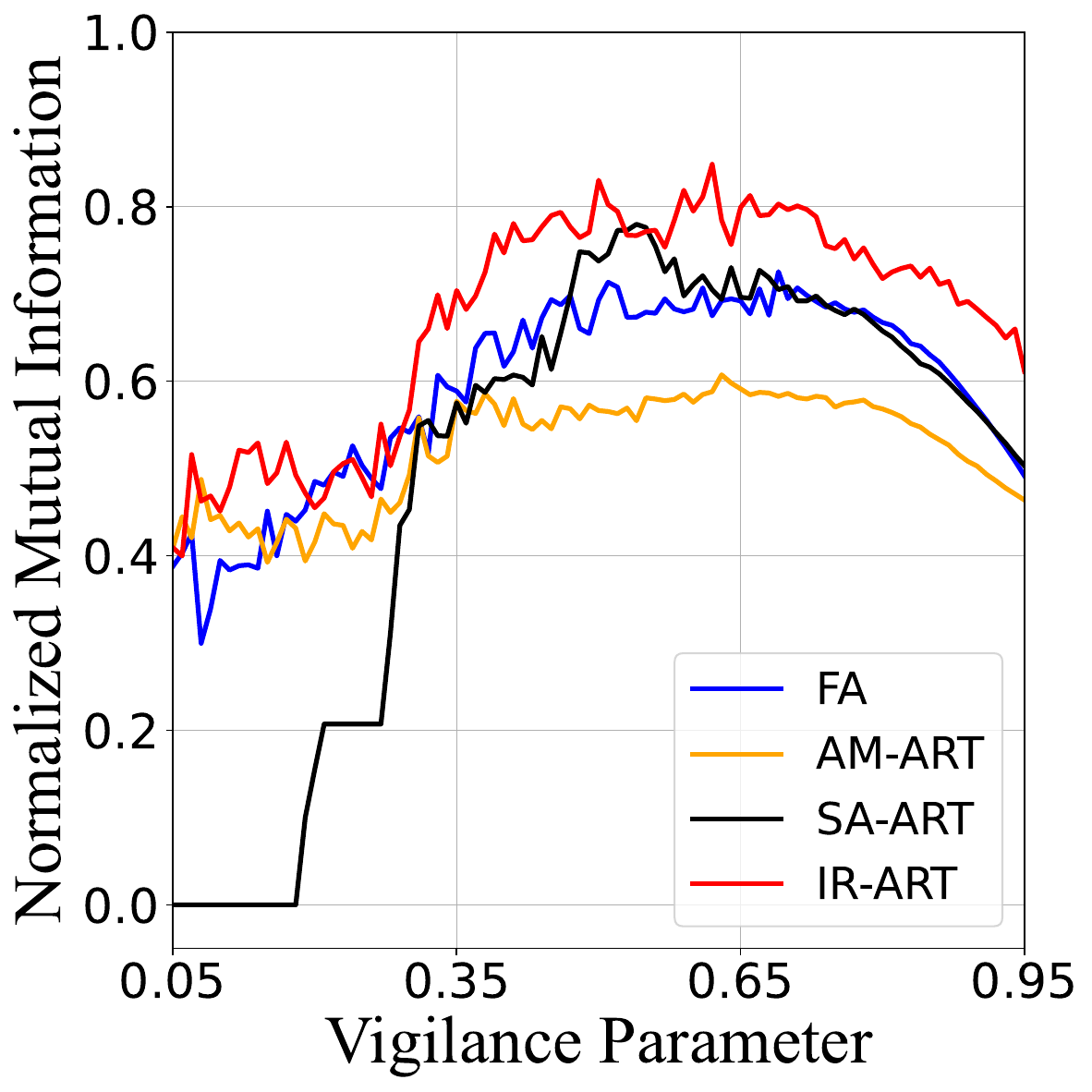} 
        \caption{Aggregation}
    \end{subfigure}\hfill
    \begin{subfigure}[b]{0.25\textwidth}
        \centering
        \includegraphics[width=\textwidth]{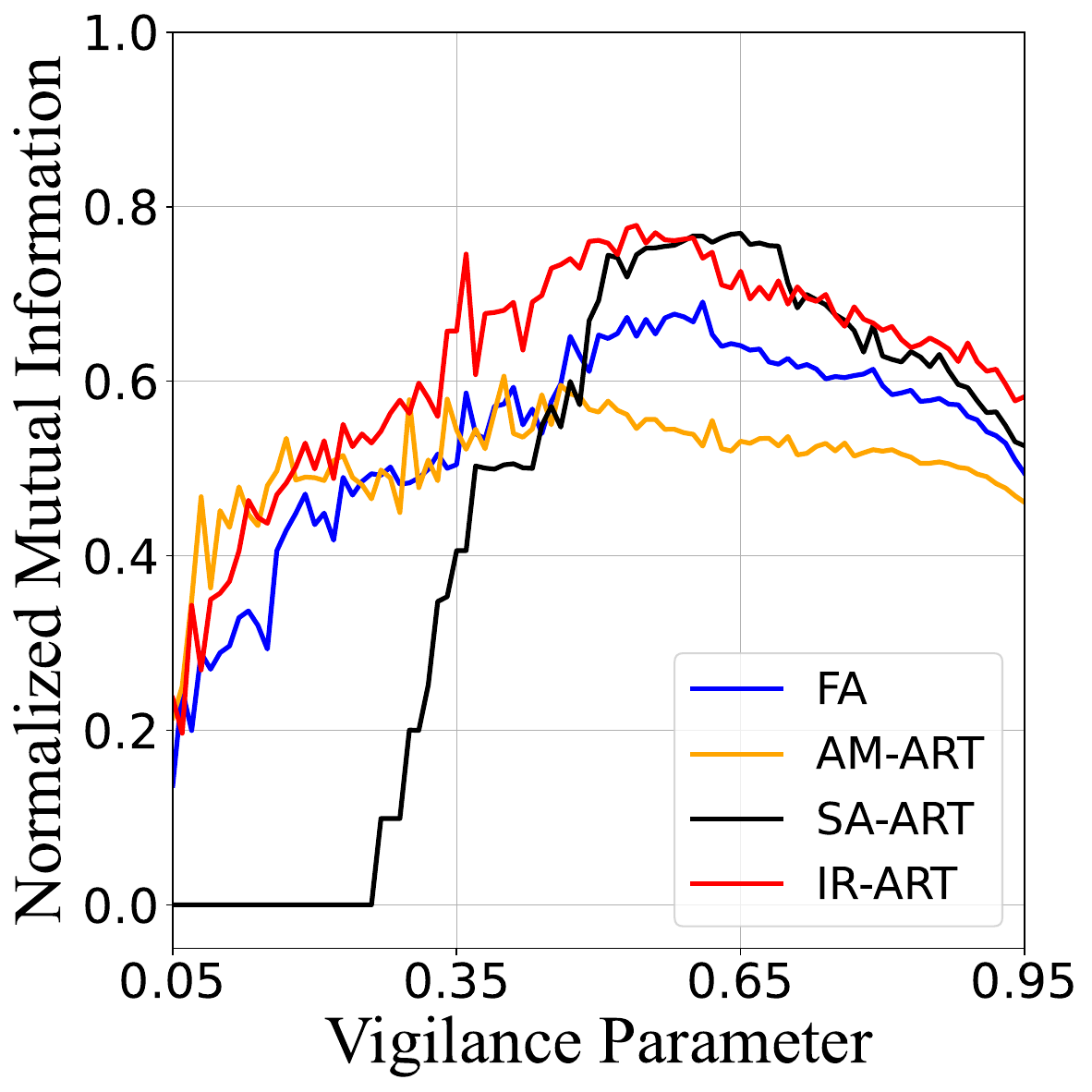} 
        \caption{Compound}
    \end{subfigure}\hfill
    \begin{subfigure}[b]{0.25\textwidth}
        \centering
        \includegraphics[width=\textwidth]{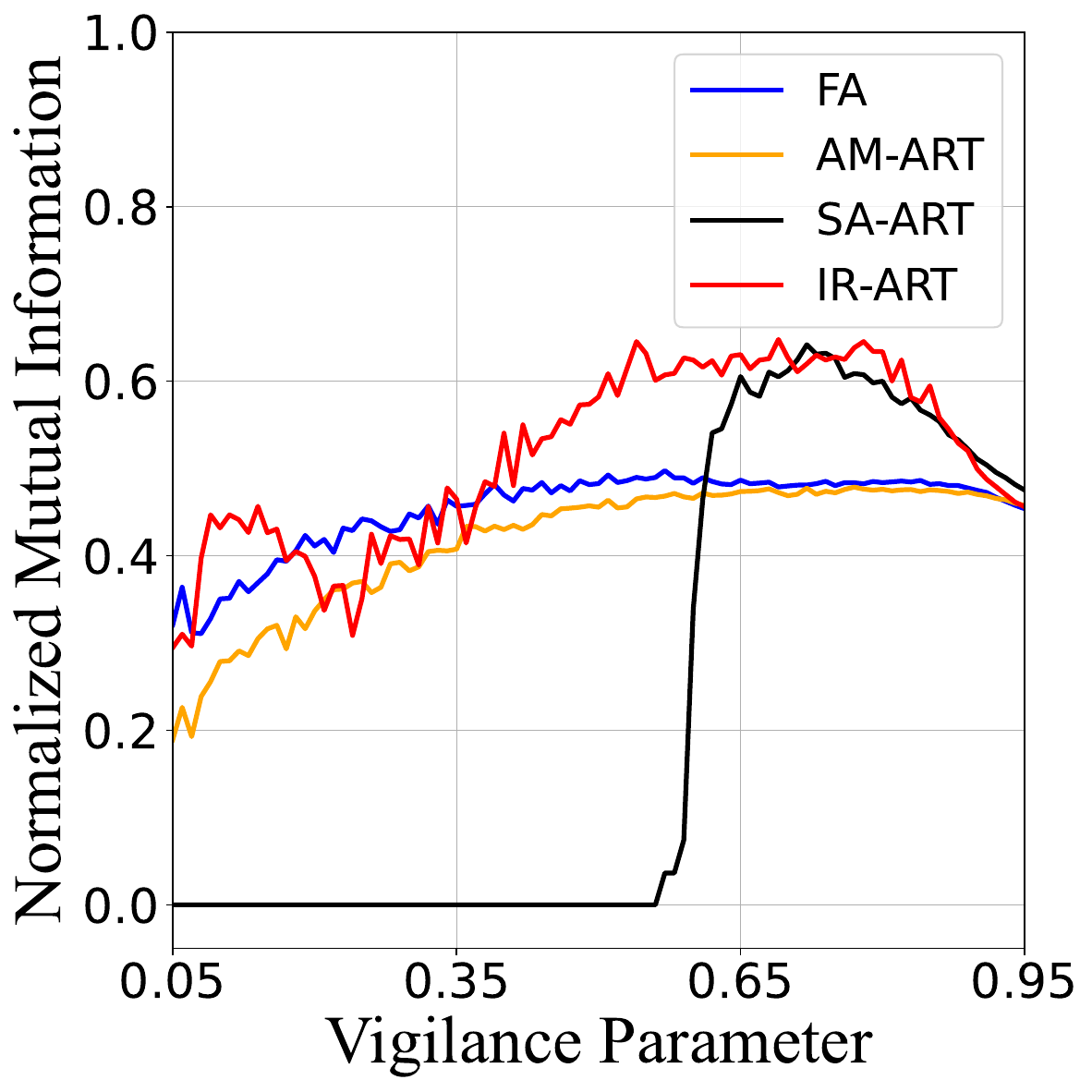} 
        \caption{Dermatology}
    \end{subfigure}\hfill
    \begin{subfigure}[b]{0.25\textwidth}
        \centering
        \includegraphics[width=\textwidth]{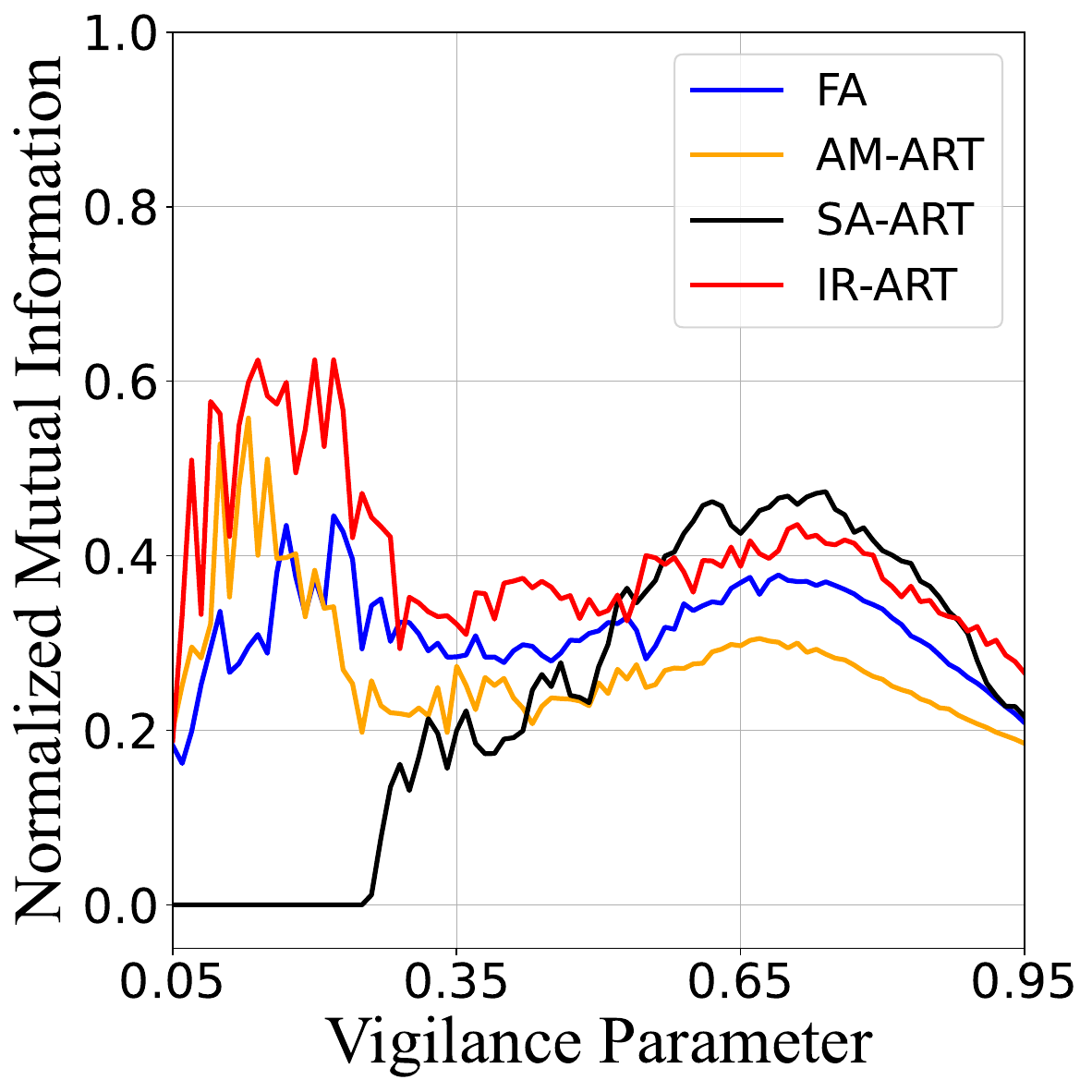} 
        \caption{Face}
    \end{subfigure}
    \\
    \begin{subfigure}[b]{0.25\textwidth}
        \centering
        \includegraphics[width=\textwidth]{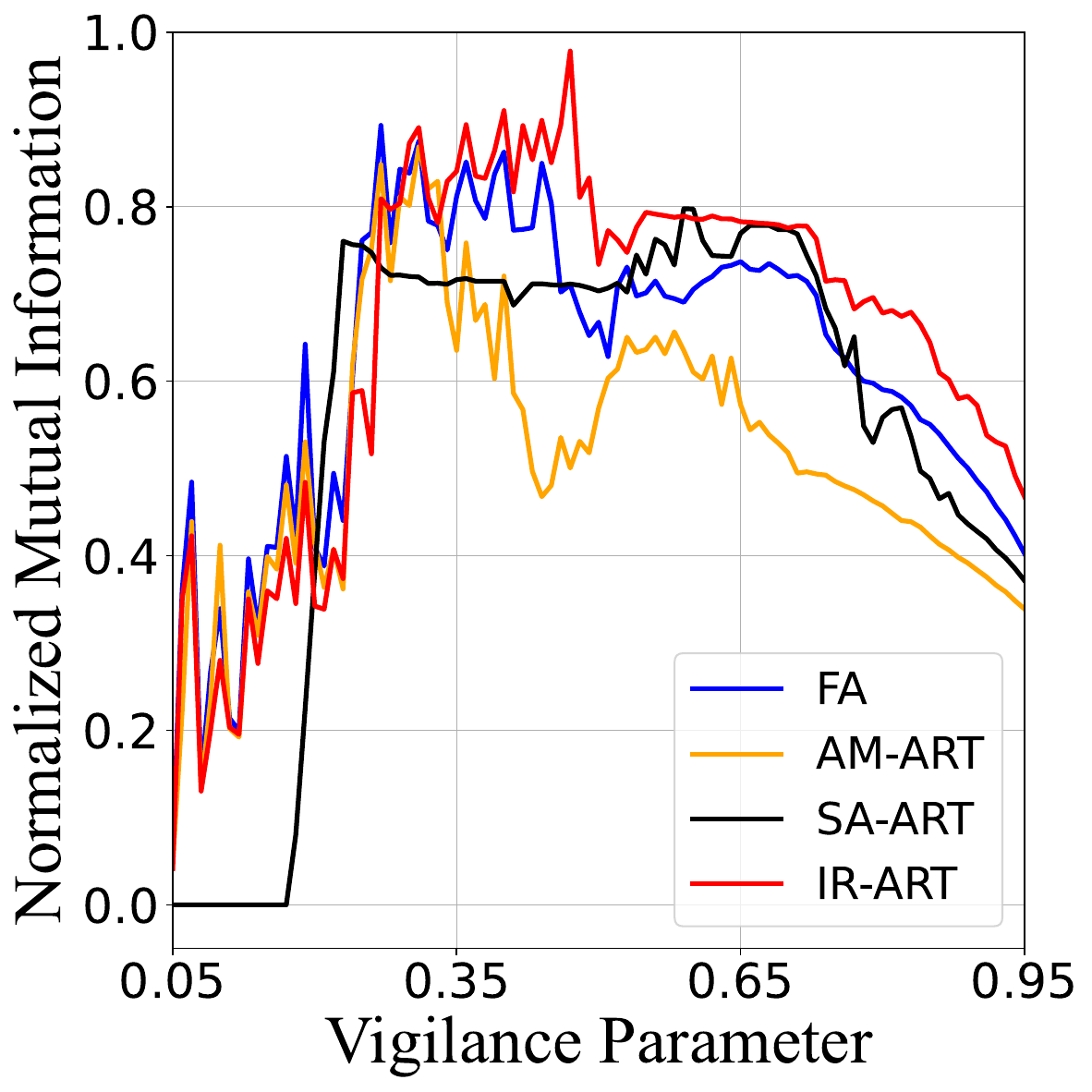}
        \caption{Flag}
    \end{subfigure}\hfill
    \begin{subfigure}[b]{0.25\textwidth}
        \centering
        \includegraphics[width=\textwidth]{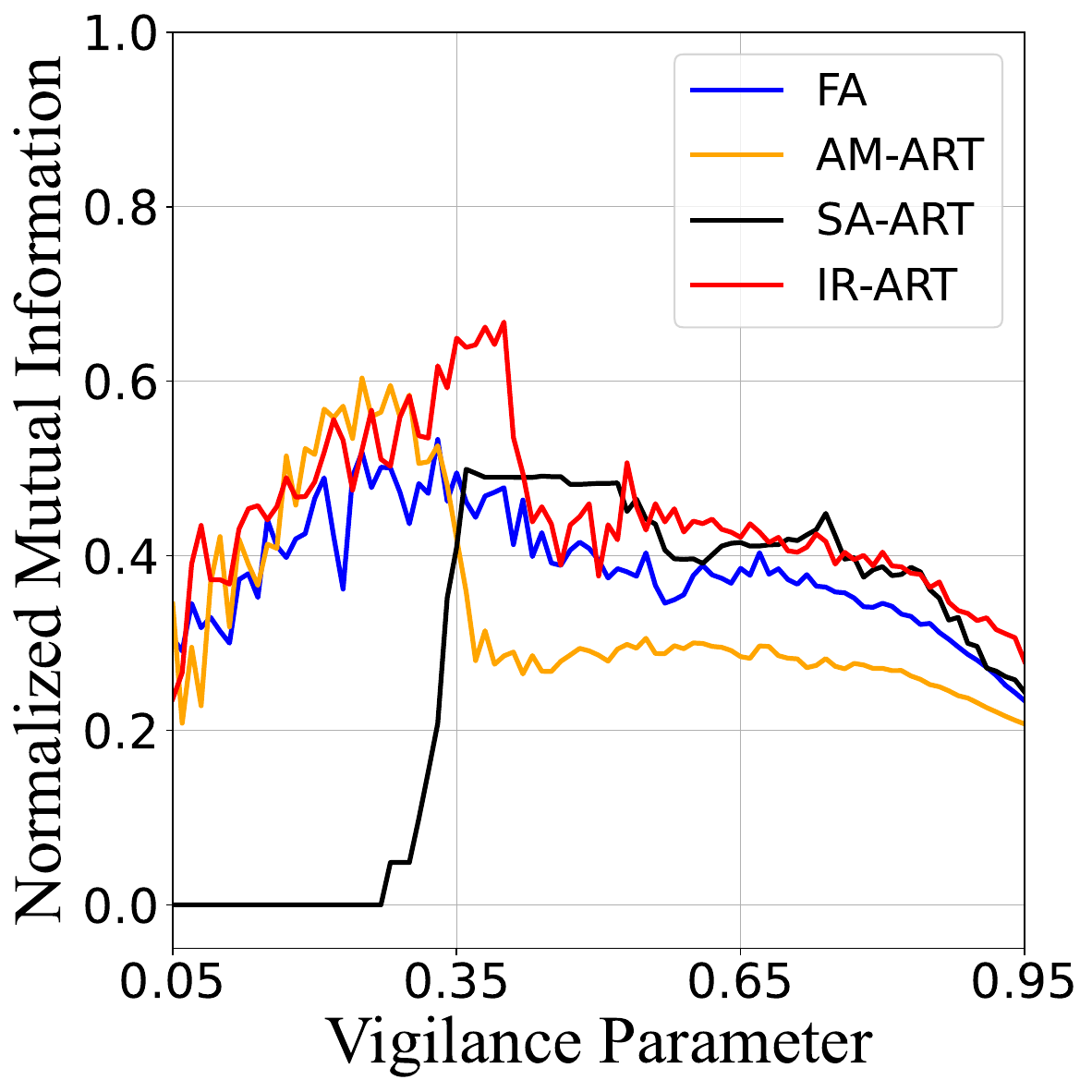} 
        \caption{Jain}
    \end{subfigure}\hfill
    \begin{subfigure}[b]{0.25\textwidth}
        \centering
        \includegraphics[width=\textwidth]{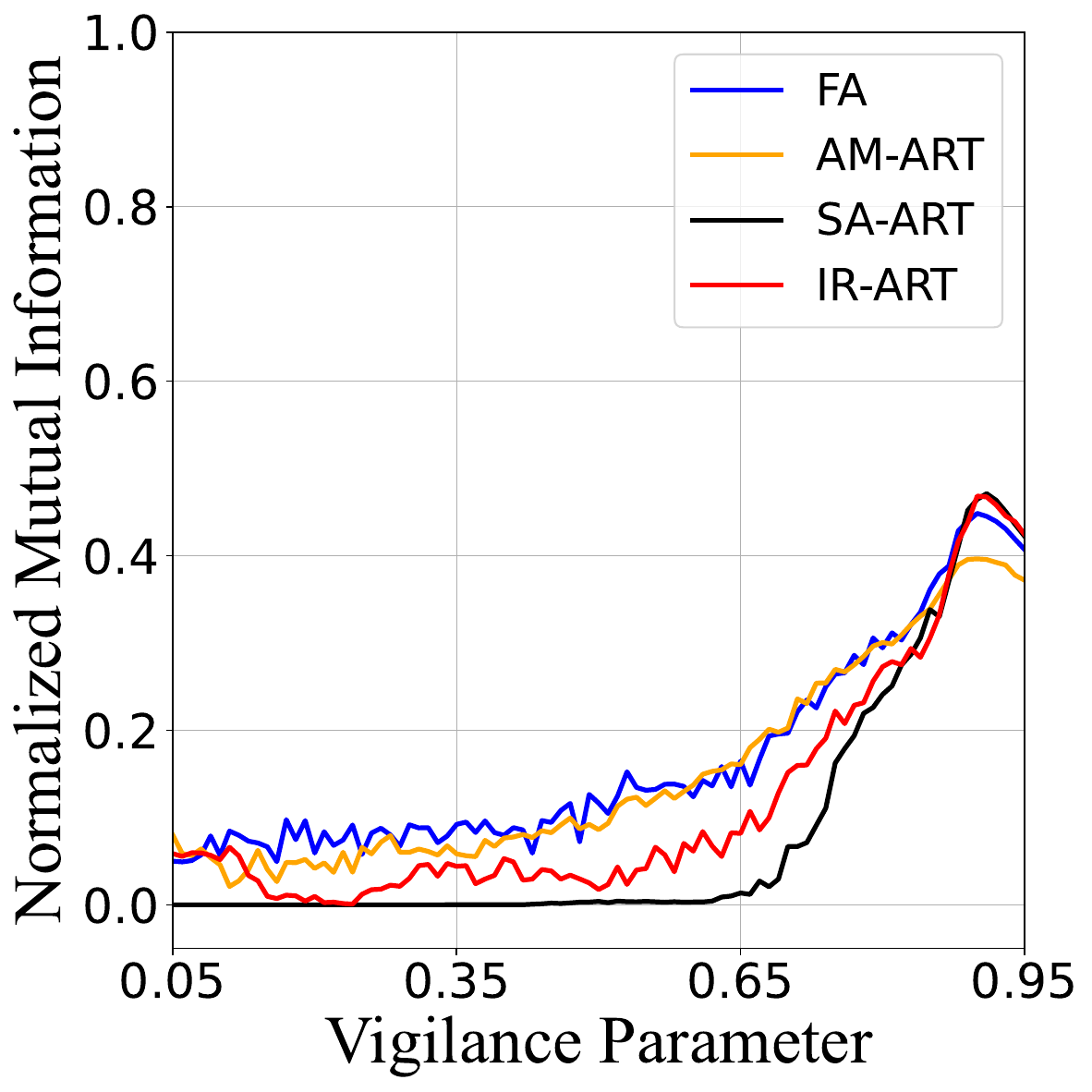}
        \caption{Spiral}
    \end{subfigure}\hfill
    \begin{subfigure}[b]{0.25\textwidth}
        \centering
        \includegraphics[width=\textwidth]{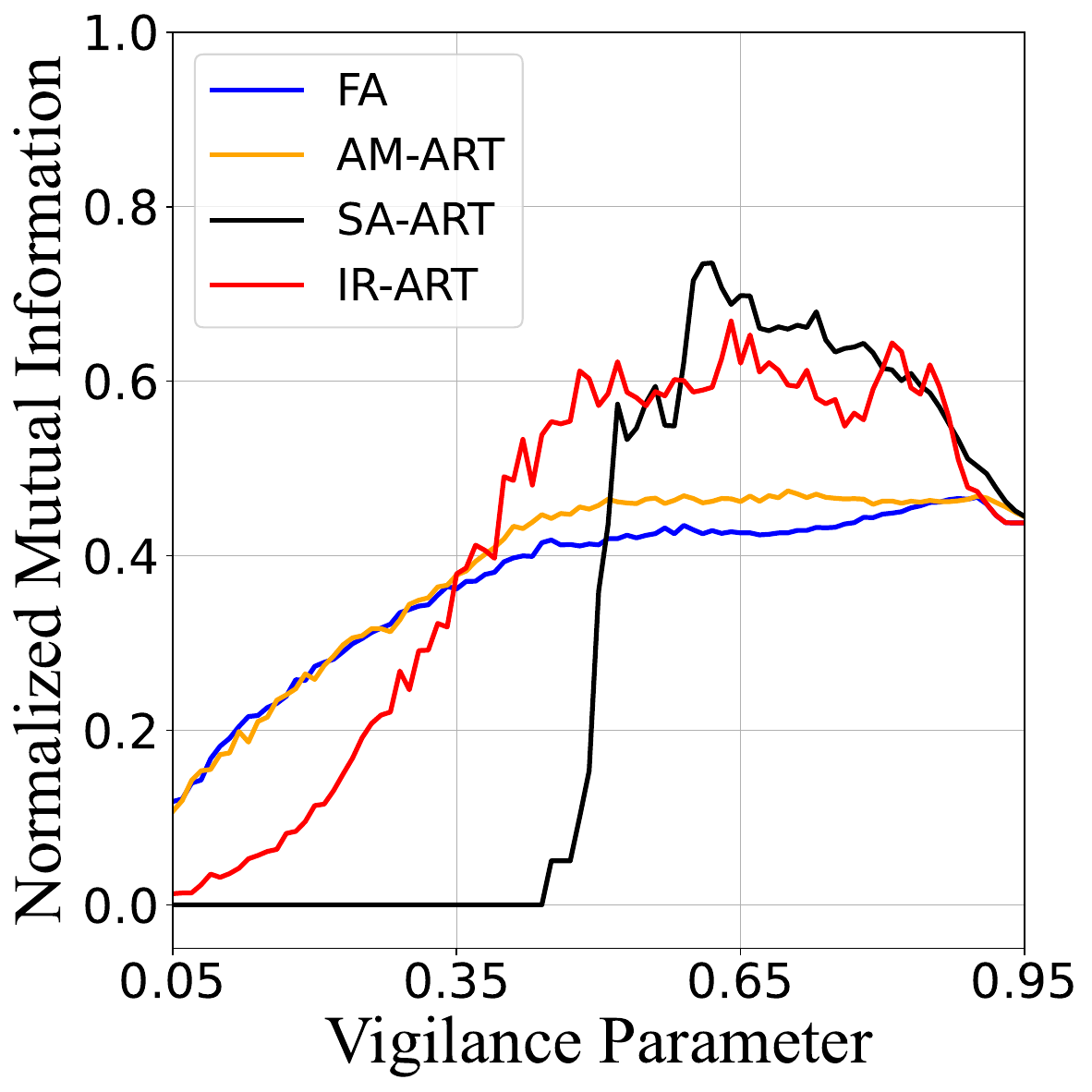} 
        \caption{Synthetic Control}
    \end{subfigure}
    
    \caption{Average NMI during $\rho$ scan on eight datasets for Fuzzy ART, AM-ART, SA-ART, and IR-ART.}
    \label{fig:index_line_chart}
    \vspace{-0.5cm}
\end{figure*}

\begin{figure}[tbp] 
    \centering
    \begin{subfigure}[b]{0.5\linewidth}
        \centering
        \includegraphics[width=\textwidth]{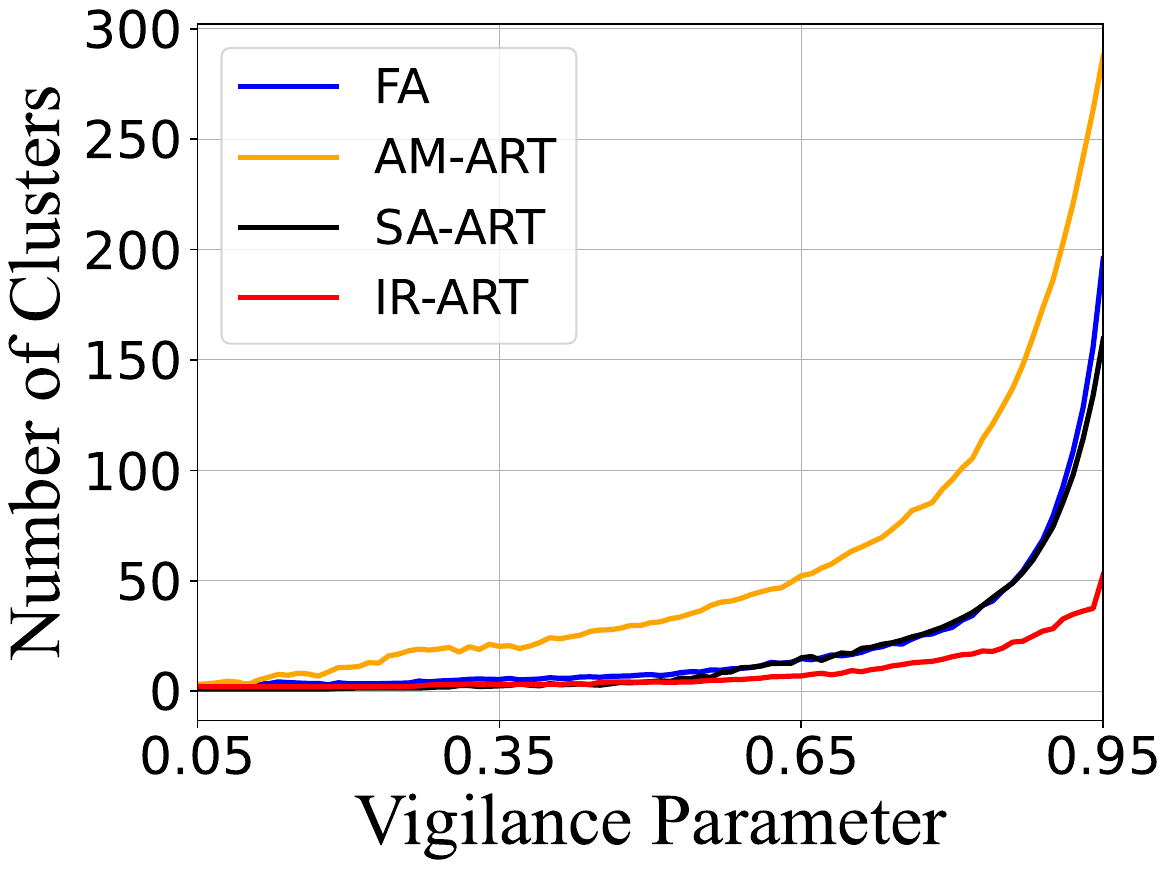}
        \caption{Aggregation}
    \end{subfigure}\hfill
    \begin{subfigure}[b]{0.5\linewidth}
        \centering
        \includegraphics[width=\textwidth]{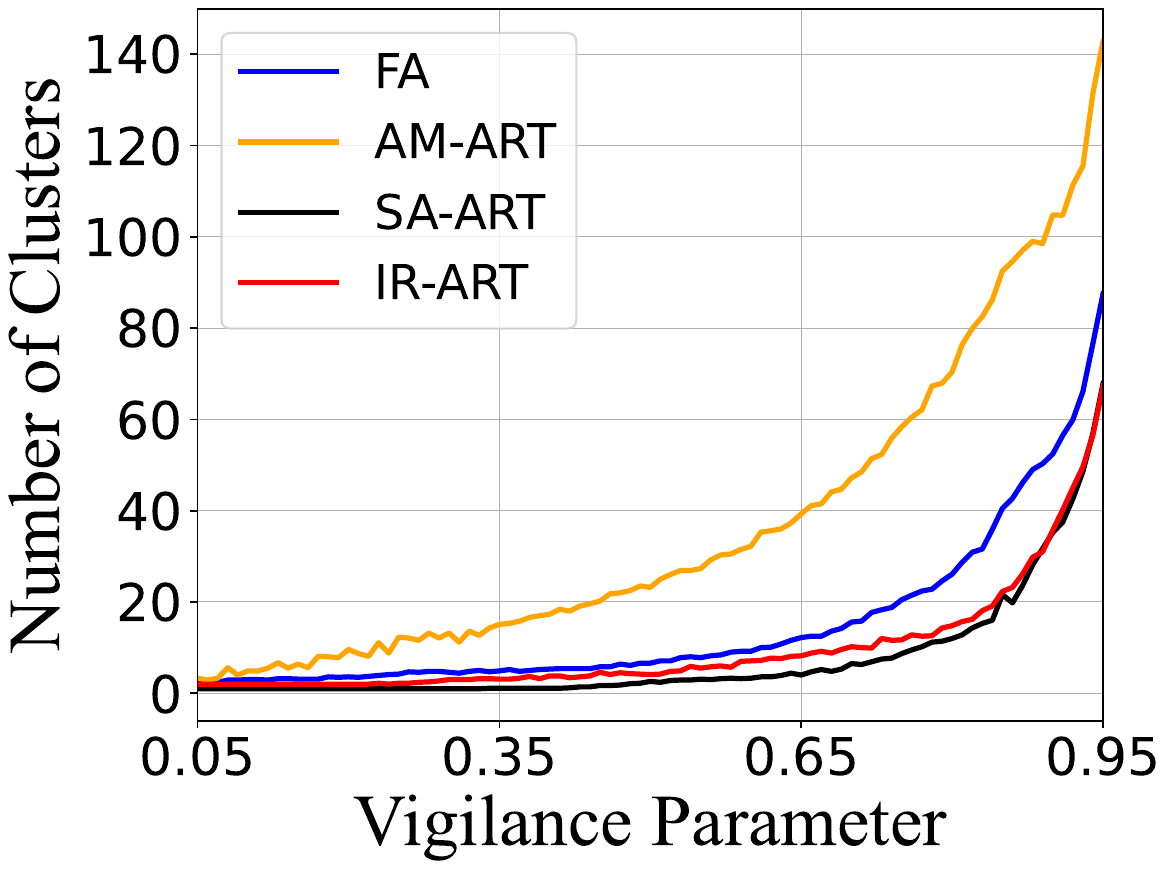}
        \caption{Spiral}
    \end{subfigure}
    \caption{Average number of clusters during $\rho$ scan.}
    \label{fig:number_of_clusters}
    \vspace{-0.5cm}
\end{figure}

To further validate the robustness of IR-ART to $\rho$, we selected eight representative datasets and plotted the complete variation of aNMI, as shown in Fig.~\ref{fig:index_line_chart}. We plotted only four methods with multiple iterations for clarity, omitting the less effective CM-ART and HI-ART that perform a single iteration. It can be seen that, in the absence of prior knowledge, poor parameter settings can lead to AM-ART performing worse than Fuzzy ART, as indicated in Fig.~\ref{fig:index_line_chart}(a). SA-ART achieves the best peak performance on a few datasets such as Fig.~\ref{fig:index_line_chart}(h), demonstrating its advantage in dealing with high-dimensional data. However, it exhibits an extreme sensitivity to $\rho$ on all datasets, with suboptimal values that significantly reduce its Mean Performance and increase its volatility. On most datasets, IR-ART significantly outperforms Fuzzy ART, regardless of whether \( \rho \) is optimal, and exhibits a smaller fluctuation in performance than SA-ART as \( \rho \) changes. These findings indicate that IR-ART can generally enhance the performance of Fuzzy ART throughout the scanning process, showing notable robustness to changes in \( \rho \), making it a suitable choice when prior knowledge is limited. However, on some datasets with special cluster shapes, IR-ART performs poorly when $\rho$ is not optimal, even performing worse than Fuzzy ART, as shown in Fig.~\ref{fig:index_line_chart}(g). This issue may be related to the excessive deletion of clusters.

Fig.~\ref{fig:number_of_clusters} shows the average number of clusters obtained during the \( \rho \) scanning process on two datasets. It can be seen that IR-ART produces a relatively small number of clusters due to its deletion of unreliable clusters. However, this may also lead to excessive cluster deletion, which could adversely affect clustering performance in some cases.

\subsection{Case Study}
\begin{figure*}[tbp]
    \centering
    \begin{subfigure}[b]{0.25\textwidth}
        \centering
        \includegraphics[width=\textwidth]{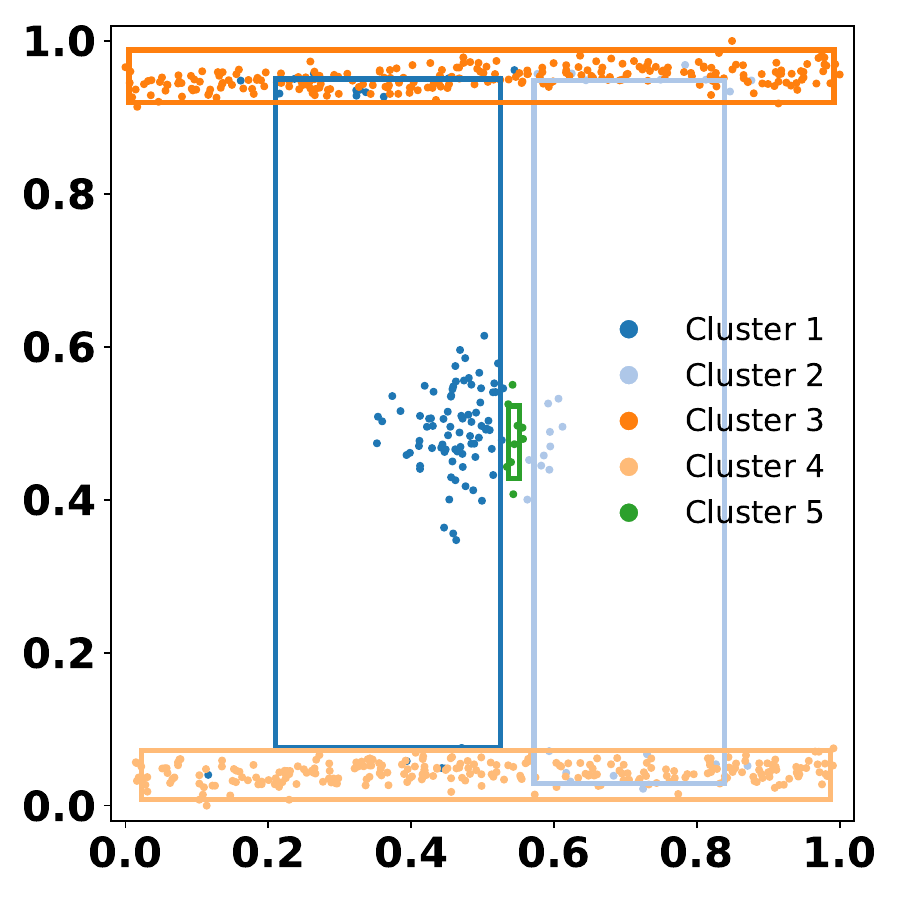} 
        \caption{Iteration 1}
    \end{subfigure}\hfill
    \begin{subfigure}[b]{0.25\textwidth}
        \centering
        \includegraphics[width=\textwidth]{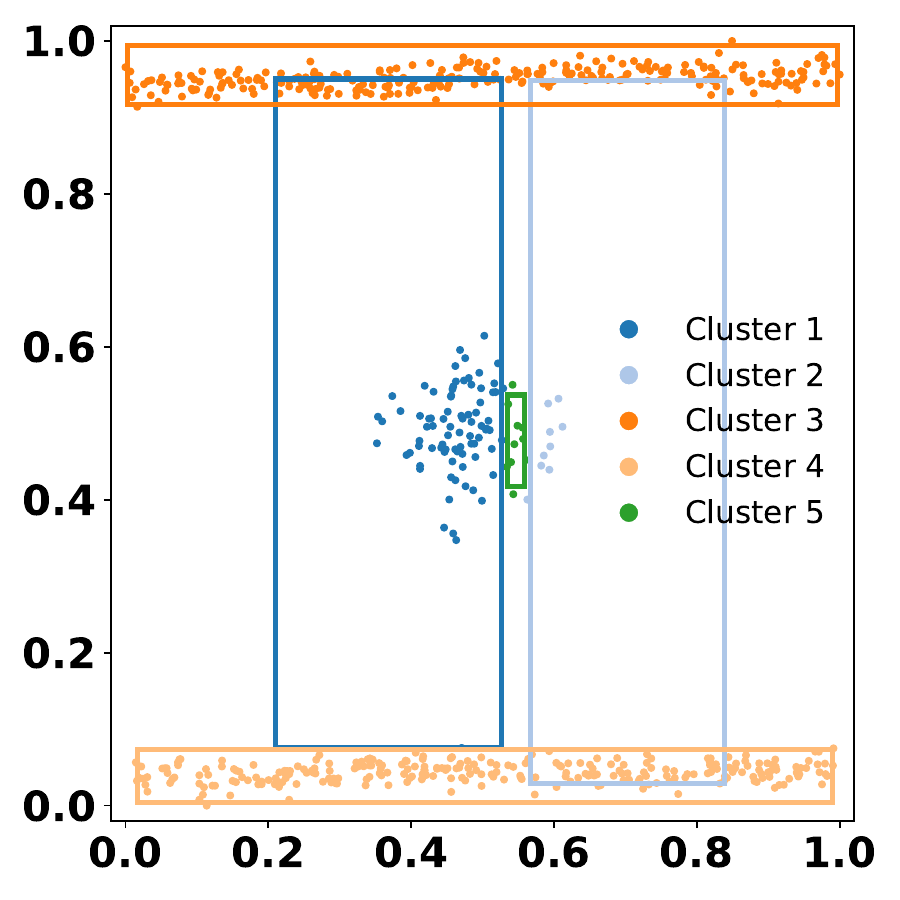} 
        \caption{Iteration 2}
    \end{subfigure}\hfill
    \begin{subfigure}[b]{0.25\textwidth}
        \centering
        \includegraphics[width=\textwidth]{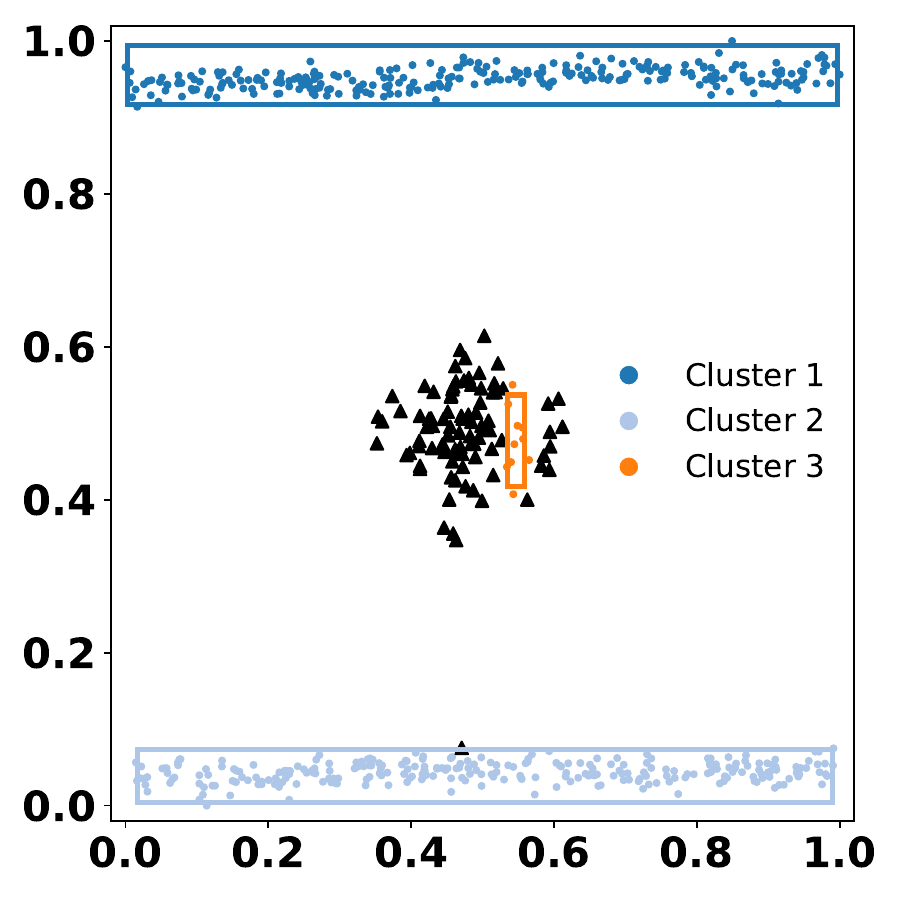} 
        \caption{Iteration 2}
    \end{subfigure}\hfill
    \begin{subfigure}[b]{0.25\textwidth}
        \centering
        \includegraphics[width=\textwidth]{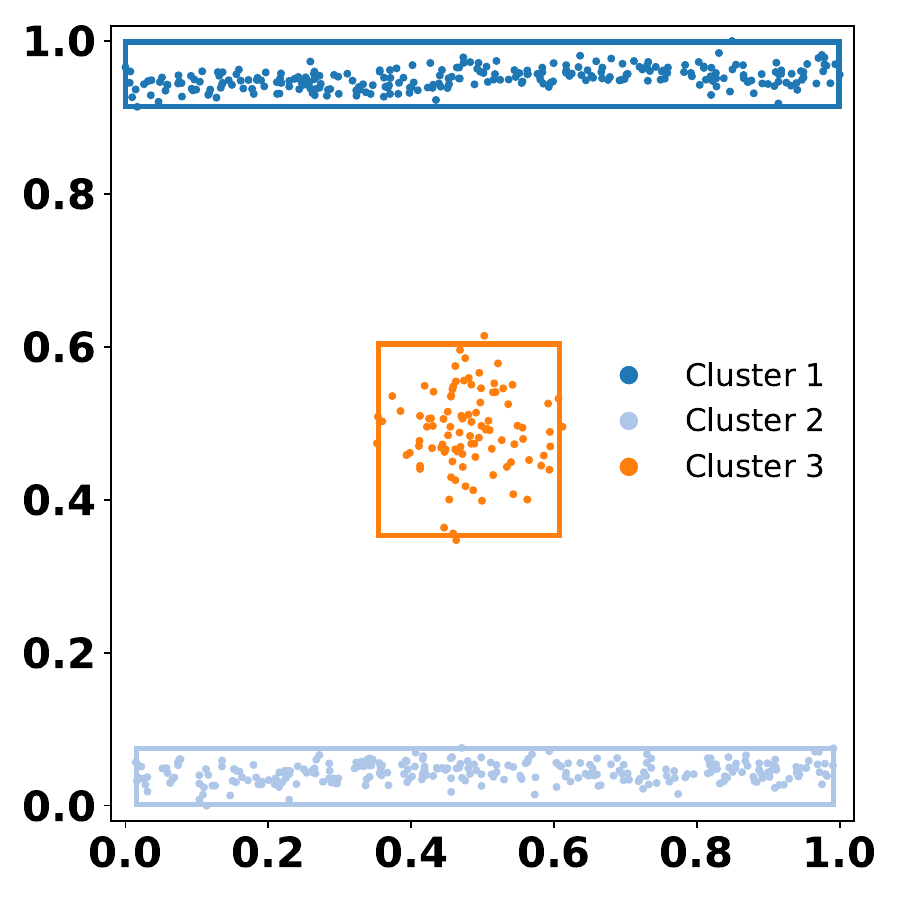} 
        \caption{Iteration 3}
    \end{subfigure}

    \caption{Clustering results on Flag dataset during the IR-ART's iteration process.}
    \label{fig:case_study}
    \vspace{-0.5cm}
\end{figure*}

As shown in Fig.~\ref{fig:case_study}, we present the clustering results of IR-ART on the Flag dataset with a random input order and \(\rho = 0.4\), using the same parameter settings as described in the Experimental Setup. Samples and cluster weight rectangles are visualized in the feature space. Fig.~\ref{fig:case_study}(a) shows the results after one execution of Fuzzy ART in Iteration 1, where cluster overlap occurs. In Fig.~\ref{fig:case_study}(b), Clusters 1 and 2 exhibit a reduced sample size after Iteration 2, and are subsequently deleted during the CSD and UCD phases, as shown in Fig.~\ref{fig:case_study}(c). The black triangles represent unassigned samples following cluster deletion. Fig.~\ref{fig:case_study}(d) illustrates that, after one execution of Fuzzy ART in Iteration 3, these samples are reassigned, and the termination criteria are met, ultimately achieving perfect clustering results on the Flag dataset. We also found that values of \(\rho\) within the range of [0.27, 0.46] (with the input order affecting this range) yield the same optimal results, indicating that IR-ART exhibits robust performance against variations in \(\rho\), thus reducing the user's reliance on expertise regarding \(\rho\).

\section{Conclusion}

\noindent Fuzzy ART is highly sensitive to the vigilance parameter, which places significant demands on prior knowledge. To address this issue, we optimize the iterative process of Fuzzy ART and propose Iterative Refinement Fuzzy Adaptive Resonance Theory (IR-ART). IR-ART integrates three key phases into the iterative process, continuously removing unstable clusters and fine-tuning the vigilance parameter.
Our experimental results show that IR-ART consistently outperforms traditional Fuzzy ART on various datasets, even with suboptimal $\rho$. This indicates that IR-ART exhibits strong robustness to \(\rho\) and can be effectively applied in scenarios where clustering needs to be performed with limited expertise. Additionally, It retains the simplicity of Fuzzy ART without adding preset parameters, making it more practical for users with limited expertise.

However, our experiments revealed some limitations of IR-ART, such as repetitive loops, excessive vigilance region expansion, and difficulties with certain cluster shapes. These issues highlight opportunities for improvement. Moreover, IR-ART can serve as a general framework that may be integrated with other ART-based clustering algorithms, offering promising directions for future research. In particular, its adaptability suggests strong potential for further application in domains such as federated learning \cite{wang2023dafkd,wang2024feddse,FedSSA,yi2024federated,liu2023cross,qi2023cross} and recommendation algorithms \cite{ma2024multimodal,ma2024plug,ma2024seedrec}.

\section{Acknowledgments}
This work is supported in part by the Shandong Province
Excellent Young Scientists Fund Program (Overseas) (Grant
no. 2022HWYQ-048), and the Shandong Province Youth Entrepreneurship Technology Support Program for Higher Education Institutions, 2022KJN028.

\bibliographystyle{IEEEtran}
\bibliography{bib/IEEEabrv,bib/ref}

\end{document}